\documentclass[letterpaper,10pt,conference]{ieeeconf}
\IEEEoverridecommandlockouts
\usepackage{graphicx} 
\usepackage{booktabs}
\usepackage{amsfonts,amsmath,amssymb}
\usepackage{tikz}
\usetikzlibrary{arrows.meta,positioning}
\usetikzlibrary{calc}

\usepackage{hyperref}
\title{A Probabilistic, Rotation-Preserving Kernel Operator Surrogate for Continuum-Robot Design Spaces}
\author{
    Branden Frieden$^{1}$,
    Alan Kuntz$^{2}$, and 
    Varun Shankar$^{1}$
\thanks{$^{1}$The Kahlert School of Computing at the University of Utah, Salt Lake City, UT 84112, USA.}
\thanks{$^{2}$The Departments of Computer Science and Electrical and Computer Engineering at Vanderbilt University, Nashville, TN, 37203, USA.}
\thanks{VS was supported by NSF DMS 2505986. Research reported in this publication was also supported by the Advanced Research Projects Agency for Health (ARPA-H) under Award Number D24AC00415-00 for ALISS. The ARPA-H award provided 25\% of total costs with an award total of up to \$10,741,534.20. The content is solely the responsibility of the authors and does not necessarily represent the official views of \mbox{ARPA-H}.}
}

\begin{document}

\maketitle

\begin{abstract}
We develop an uncertainty-aware, operator-learning-based surrogate for
tendon-driven continuum robots (TDCRs) that generalizes across robot design
and actuation while predicting the complete equilibrium configuration.
Our central idea is a nonstationary Gaussian-process (GP) kernel whose metric
is learned by a shallow neural network from mechanics-aware design
coordinates. The network learns how similarity changes across design space,
while the resulting kernel defines the design-to-pose operator, its posterior
uncertainty, and a criterion for sequential simulator acquisition. We represent
backbone position and rotational strain with spatial kernels and reconstruct
orientation directly on $SO(3)$, preserving valid material frames and tendon
geometry. We use ASkotch with Nystr\"om preconditioning for scalable kernel operator
solves and a separate Nystr\"om approximation for posterior covariance.
Experiments show that kernel-coverage acquisition reduces mean trajectory
error by approximately $15\%$ relative to IID sampling, while posterior
variance increases for increasingly unsupported designs. Compared with prior
neural-operator TDCR models, the method is 30--50\% more accurate using
approximately $5\times$ fewer training designs while additionally providing
posterior uncertainty estimates.
\end{abstract}

\section{Introduction}
\label{sec:intro}

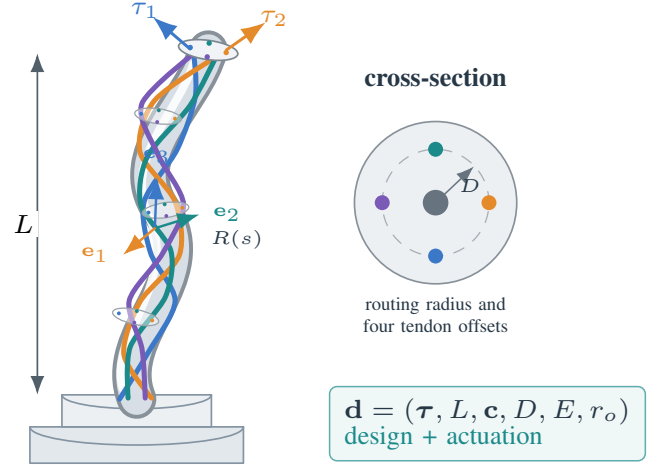
\begin{figure}[t]
    \centering
    \resizebox{0.5\textwidth}{!}{
\begin{tikzpicture}[x=1cm,y=1cm,>=Latex,line cap=round,line join=round]
  \definecolor{uTeal}{HTML}{1B8A89}
  \definecolor{uBlue}{HTML}{3B78C8}
  \definecolor{uPurple}{HTML}{7A5AB7}
  \definecolor{uOrange}{HTML}{E58A2B}
  \definecolor{uGold}{HTML}{D6A326}
  \definecolor{uInk}{HTML}{263746}
  \definecolor{uGray}{HTML}{D7DEE5}
  \definecolor{uGrayDark}{HTML}{8494A3}

  \fill[uGray!70] (1.55,0.18) rectangle (3.55,0.52);
  \draw[uGrayDark,line width=.45pt] (1.55,0.18) rectangle (3.55,0.52);
  \fill[uGray!45] (1.85,0.52) rectangle (3.25,0.82);
  \draw[uGrayDark,line width=.45pt] (1.85,0.52) rectangle (3.25,0.82);
  \draw[uGrayDark,line width=.45pt] (1.55,0.52) arc[start angle=180,end angle=360,x radius=1.0,y radius=.17];
  \draw[uGrayDark,line width=.45pt] (1.85,0.82) arc[start angle=180,end angle=360,x radius=.70,y radius=.13];

  \path[draw=uInk!55,line width=10.5pt]
    (2.55,.78) .. controls (2.30,1.40) and (3.20,1.90) .. (2.78,2.55)
                .. controls (2.42,3.08) and (3.08,3.55) .. (3.22,4.05);
  \path[draw=uGray,line width=8.4pt]
    (2.55,.78) .. controls (2.30,1.40) and (3.20,1.90) .. (2.78,2.55)
                .. controls (2.42,3.08) and (3.08,3.55) .. (3.22,4.05);
  \path[draw=white!85,line width=2.2pt,opacity=.9]
    (2.42,.88) .. controls (2.28,1.38) and (2.96,1.88) .. (2.66,2.50)
                .. controls (2.42,3.02) and (2.94,3.47) .. (3.08,3.92);

  \draw[uBlue,line width=1.35pt]
    plot[smooth] coordinates {(2.39,.78) (2.66,1.22) (2.91,1.70) (2.68,2.15) (2.62,2.63) (2.93,3.12) (3.10,3.62) (3.08,4.06)};
  \draw[uOrange,line width=1.35pt]
    plot[smooth] coordinates {(2.69,.79) (2.43,1.20) (2.58,1.69) (2.95,2.13) (2.91,2.63) (2.58,3.10) (2.72,3.58) (3.35,4.02)};
  \draw[uTeal,line width=1.35pt]
    plot[smooth] coordinates {(2.47,.79) (2.49,1.27) (2.82,1.74) (2.87,2.17) (2.59,2.60) (2.65,3.05) (3.05,3.53) (3.30,4.00)};
  \draw[uPurple,line width=1.35pt]
    plot[smooth] coordinates {(2.63,.79) (2.60,1.25) (2.48,1.70) (2.79,2.14) (2.99,2.60) (2.75,3.07) (2.60,3.55) (3.15,4.07)};

  \foreach \x/\y/\rot in {2.54/1.55/-12,2.82/2.55/8,2.75/3.43/-8}{
    \begin{scope}[shift={(\x,\y)},rotate=\rot]
      \draw[uInk!45,line width=.35pt,fill=white,fill opacity=.45] (0,0) ellipse[x radius=.22,y radius=.07];
      \fill[uBlue]   (-.16,0) circle (.018);
      \fill[uOrange] (.16,0) circle (.018);
      \fill[uTeal]   (0,.055) circle (.018);
      \fill[uPurple] (0,-.055) circle (.018);
    \end{scope}
  }

  \begin{scope}[shift={(3.22,4.05)},rotate=-8]
    \draw[uInk!60,fill=uGray!35,line width=.45pt] (0,0) ellipse[x radius=.28,y radius=.09];
    \fill[uBlue]   (-.17,0) circle (.025);
    \fill[uOrange] (.17,0) circle (.025);
    \fill[uTeal]   (0,.065) circle (.025);
    \fill[uPurple] (0,-.065) circle (.025);
  \end{scope}
  \draw[-Latex,uBlue,line width=.8pt]   (3.05,4.06)--(2.70,4.36);
  \draw[-Latex,uOrange,line width=.8pt] (3.39,4.05)--(3.76,4.31);
  \node[font=\scriptsize,text=uBlue] at (2.63,4.42) {$\tau_1$};
  \node[font=\scriptsize,text=uOrange] at (3.84,4.35) {$\tau_2$};

  \draw[<->,uInk!80,line width=.45pt] (1.62,.82) -- (1.62,4.02);
  \node[font=\scriptsize,fill=white,inner sep=1pt] at (1.49,2.42) {$L$};

  \coordinate (F) at (2.72,2.38);
  \draw[-Latex,uBlue,line width=.65pt]   (F)--++(.02,.48) node[above,font=\tiny] {$\mathbf e_3$};
  \draw[-Latex,uTeal,line width=.65pt]   (F)--++(.42,.13) node[right,font=\tiny] {$\mathbf e_2$};
  \draw[-Latex,uOrange,line width=.65pt] (F)--++(-.30,-.25) node[left,font=\tiny] {$\mathbf e_1$};
  \node[font=\tiny,text=uInk,anchor=west] at (3.12,2.28) {$R(s)$};

  \begin{scope}[shift={(5.35,2.62)}]
    \node[font=\scriptsize\bfseries,text=uInk] at (0,1.18) {cross-section};
    \fill[uGray!45] (0,0) circle (.76);
    \draw[uInk!55,line width=.45pt] (0,0) circle (.76);
    \draw[uInk!40,dashed,line width=.35pt] (0,0) circle (.50);
    \fill[uInk!70] (0,0) circle (.12);
    \fill[uBlue]   (0,-.50) circle (.07);
    \fill[uOrange] (.50,0) circle (.07);
    \fill[uTeal]   (0,.50) circle (.07);
    \fill[uPurple] (-.50,0) circle (.07);
    \draw[-Latex,uInk!70,line width=.4pt] (0,0)--(.36,.34);
    \node[font=\tiny,text=uInk] at (.32,.16) {$D$};
    \node[font=\tiny,text=uInk,align=center] at (0,-1.05) {routing radius and\\four tendon offsets};
  \end{scope}

  \node[anchor=west,rounded corners=2pt,draw=uTeal!70,fill=uTeal!7,inner xsep=4pt,inner ysep=3pt,
        font=\scriptsize,text=uInk,align=left] at (4.35,.55)
        {$\mathbf d=(\boldsymbol\tau,L,\mathbf c,D,E,r_o)$\\[-1pt]\textcolor{uTeal}{design + actuation}};

  \path[use as bounding box] (1.15,.05) rectangle (6.35,4.55);
\end{tikzpicture}}
    \caption{Four-tendon TDCR and the continuous configuration learned by the
    surrogate. The input contains robot-design and actuation parameters; the
    output is the backbone/material-frame pose and tendon trajectories.}
    \label{fig:tdcr}
\end{figure}

A continuum-robot surrogate becomes substantially more useful when it spans
\emph{robot design} as well as actuation, returns the complete equilibrium
shape, and indicates when a query is poorly supported by data. Tendon-driven
continuum robots (TDCRs) are attractive for minimally invasive surgery and
confined-space manipulation because of their compact actuation and large,
smooth deformations~\cite{Webster2010ConstantCurvature,Burgner2015_TRO,
Dupont2022_ProcIEEE}. Geometrically exact Cosserat-rod models capture tendon
loading, backbone mechanics, and routing geometry
~\cite{Rucker2011StaticsTendons,Rao2021HowToModelTDCR}, but repeated
boundary-value solves become costly in planning, control, uncertainty analysis,
and design optimization~\cite{Till2019RealtimeCosserat}. In this work, we
focus on operator-learning-based, uncertainty-aware surrogate models for TDCRs
that take design and actuation parameters as inputs and supply full
backbone/material-frame pose and routed tendon trajectories as outputs, together
with posterior uncertainty estimates; see Figure ~\ref{fig:tdcr}.

Operator learning has recently been leveraged in several robotics applications.
Examples include DeepONet-type inverse load reconstruction for Cosserat rods
~\cite{Tang2025HapticNO}, neural operators for full-shape soft-robot inverse
kinematics~\cite{VeilFlaschelKuhlDellaSantina2026}, and design-space neural
operators for TDCR equilibria~\cite{FriedenFergusonKuntzShankar2026}.
However, neural operators do not typically come with uncertainty estimates,
requiring reformulation, ensembles, or post-hoc modifications
~\cite{YangKissasPerdikaris2022,LinMoyaZhang2023,GargChakraborty2023,
MaPittAzizzadenesheliAnandkumar2024,MoyaMollaaliZhangLuLin2025,
MagnaniPfortnerWeberHennig2025}. Gaussian processes (GPs) offer an inherently uncertainty-aware alternative, with growing interesting on GP-based operator-learning methods for PDEs~\cite{BatlleDarcyHosseiniOwhadi2024,KumarNayekChakraborty2024,MoraYousefpourHosseinmardiOwhadiBostanabad2025,
KumarNayekChakraborty2025}. GPs also provide uncertainty-aware models for
continuum-robot pose and strain estimation on $SE(3)$
~\cite{LilgeBarfootBurgnerKahrs2022}, forward and inverse kinematics
~\cite{RelanoMunozMonje2023}, and online probabilistic control
~\cite{MoWeiKongZhaiLiuSun2025}; more broadly, GP uncertainty supports
data-efficient robot learning~\cite{DeisenrothFoxRasmussen2015}. These
approaches primarily learn state, kinematic, or control maps for a fixed robot.

In this work, we present a probabilistic kernel/GP surrogate model for TDCRs
whose \emph{metric is learned by a neural network}. We use a small shallow neural network maps mechanics-aware coordinates to local blockwise kernel shapes, which define a positive-definite
nonstationary Mat\'ern covariance kernel~\cite{PaciorekSchervish2004}. The
network never predicts the robot pose: it learns how similarity changes across
design space. The resulting kernel simultaneously defines the design-to-pose
operator, its GP uncertainty, and a kernel-coverage criterion for acquiring new
simulator data.

We also build the output representation to respect the mechanics. Spatial kernel
coefficients represent dimensionless backbone position and rotational strain;
we integrate the predicted strain to reconstruct orientation directly on
$SO(3)$, then recover tendon paths analytically from the routing geometry. We
focus on four-tendon TDCRs and enforce the corresponding symmetry in both the
learned metric and prediction. For scalability, we use
ASkotch~\cite{askotch2026} with a Nystr\"om preconditioner to solve the kernel
systems associated with the GP mean, and we use an independent Nystr\"om
construction to approximate the posterior covariance. We then propagate both
design-space and spatial uncertainties to Cartesian covariance tubes around the
TDCR.

Our experimental results show that kernel-coverage acquisition reduces mean
trajectory error by approximately $15\%$ relative to IID acquisition over the
tested training sizes. We also find that the GP posterior variance increases
systematically as designs move away from the training distribution. The overall
method is 30--50\% more accurate, using approximately $5\times$ fewer training
designs, than prior neural-operator work for TDCRs~\cite{FriedenFergusonKuntzShankar2026}, while also providing posterior
uncertainty estimates.

The remainder of the paper is organized as follows.
Section~\ref{sec:kernel_models} develops the kernel operator surrogate,
including the rotation-preserving pose representation, mechanics-aware design
coordinates, learned nonstationary kernel metric, sequential acquisition, and
scalable mean and covariance calculations. Section~\ref{sec:results} evaluates
prediction accuracy, active acquisition, scalable kernel solves, posterior
uncertainty, and Cartesian uncertainty propagation. We conclude with the main
findings and directions for future work.
\section{Background}
\label{sec:background}

\subsection{TDCR mechanics}
\label{sec:cosserat}

We model the robot backbone as a geometrically exact Cosserat rod~\cite{Antman2005Problems,Rucker2011StaticsTendons}. Let $s\in[0,L]$ denote arclength, $\mathbf r(s)\in\mathbb R^3$ the backbone centerline, and $R(s)\in SO(3)$ the material frame. The rod kinematics are
\begin{equation}
\mathbf r'(s)=R(s)\boldsymbol\nu(s),
\qquad
R'(s)=R(s)[\boldsymbol\kappa(s)]_\times,
\label{eq:cosserat_kinematics}
\end{equation}
where $\boldsymbol\nu$ is the shear/extension strain, $\boldsymbol\kappa$ is the bending/torsion strain, and $[\cdot]_\times$ is the cross-product matrix. Internal force and moment satisfy the Cosserat balance laws and are related to these strains by constitutive matrices determined by the backbone geometry and material properties~\cite{Antman2005Problems,Rao2021HowToModelTDCR}.

For tendon $i$, let $\boldsymbol\rho_i(s)$ denote its known offset from the backbone in the local material frame. Its spatial trajectory is
\begin{equation}
\mathbf p_i(s)=\mathbf r(s)+R(s)\boldsymbol\rho_i(s).
\label{eq:tend-path}
\end{equation}
Tendon tensions enter the Cosserat equilibrium equations through distributed forces and moments induced by these routed paths~\cite{Rucker2011StaticsTendons}. Solving the resulting boundary-value problem yields $\mathbf r$, $R$, and the tendon trajectories. We use these mechanics solves as the high-fidelity data source and learn the map from design/actuation parameters to the resulting continuous configuration.

\subsection{Kernel operators and Gaussian-process uncertainty}
\label{sec:op-learn}

A design-space surrogate maps a parameter vector $\mathbf d$ to a function-valued robot configuration. Previous TDCR operator models represented this map with neural operators~\cite{FriedenFergusonKuntzShankar2026}; here we instead use kernel regression after representing the output fields in a common coefficient space. If $\mathbf z(\mathbf d)\in\mathbb R^{d_z}$ denotes those coefficients and $k_d$ is a positive-definite kernel on design space, the regularized kernel model has the form
\begin{equation}
\widehat{\mathbf z}(\mathbf d)
=\sum_{j=1}^{N} k_d(\mathbf d,\mathbf d_j)\boldsymbol\gamma_j,
\label{eq:background_kernel_operator}
\end{equation}
with coefficients obtained from one shared kernel system. Kernel constructions of this type have been shown to be competitive for operator learning~\cite{BatlleDarcyHosseiniOwhadi2024}.

The same positive-definite kernel defines a GP covariance~\cite{RasmussenWilliams2006}. Thus, the kernel fit supplies both a posterior mean and a posterior covariance measuring how strongly a query is supported by the acquired designs. We use this dual interpretation twice: a spatial kernel represents fields along normalized arclength, while a second kernel learns the design-to-coefficient map. The following section builds both kernels so that the reconstructed pose respects $SO(3)$ and the design metric reflects TDCR mechanics.

\section{Probabilistic Kernel Operator Learning}
\label{sec:kernel_models}
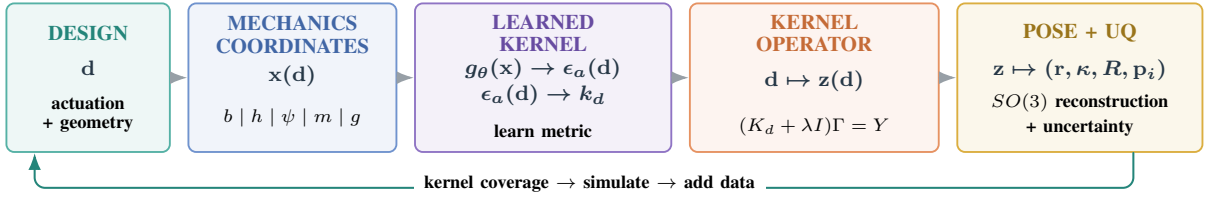
\begin{figure*}[t]
    \centering
    \resizebox{0.9\textwidth}{!}{
\begin{tikzpicture}[x=1cm,y=1cm,>=Latex,line cap=round,line join=round]
  \definecolor{uTeal}{HTML}{2A9D8F}
  \definecolor{uBlue}{HTML}{4C78C2}
  \definecolor{uPurple}{HTML}{7656B7}
  \definecolor{uOrange}{HTML}{E07A3F}
  \definecolor{uGold}{HTML}{D6A326}
  \definecolor{uInk}{HTML}{263746}

  \tikzset{
    stage/.style={rounded corners=3pt,line width=.72pt,minimum height=2.00cm,inner sep=3pt,align=center},
    head/.style={font=\footnotesize\bfseries,align=center},
    eq/.style={font=\footnotesize\bfseries,align=center,text=uInk,execute at begin node=\boldmath},
    sub/.style={font=\scriptsize\bfseries,align=center,text=black},
    flow/.style={-Latex,line width=.82pt,draw=uInk!48}
  }

  \node[stage,draw=uTeal!80,fill=uTeal!8,minimum width=2.20cm] (A) at (1.15,1.35) {};
  \node[head,text=uTeal!85!black] at ([yshift=.60cm]A.center) {DESIGN};
  \node[eq] at ([yshift=.13cm]A.center) {$\mathbf d$};
  \node[sub] at ([yshift=-.49cm]A.center) {\shortstack{actuation\\+ geometry}};

  \node[stage,draw=uBlue!80,fill=uBlue!8,minimum width=2.80cm] (B) at (3.90,1.35) {};
  \node[head,text=uBlue!85!black] at ([yshift=.58cm]B.center) {\shortstack{MECHANICS\\COORDINATES}};
  \node[eq] at ([yshift=.02cm]B.center) {$\mathbf x(\mathbf d)$};
  \node[sub] at ([yshift=-.53cm]B.center) {$b\;|\;h\;|\;\psi\;|\;m\;|\;g$};

  \node[stage,draw=uPurple!82,fill=uPurple!8,minimum width=3.45cm] (C) at (7.275,1.35) {};
  \node[head,text=uPurple!88!black] at ([yshift=.60cm]C.center) {\shortstack{LEARNED\\KERNEL}};
  \node[eq] at ([yshift=-.06cm]C.center)
    {\shortstack{$g_\theta(\mathbf x)\to\epsilon_a(\mathbf d)$\\[-1pt]$\epsilon_a(\mathbf d)\to k_d$}};
  \node[sub] at ([yshift=-.72cm]C.center) {learn metric};

  \node[stage,draw=uOrange!82,fill=uOrange!8,minimum width=3.35cm] (D) at (10.925,1.35) {};
  \node[head,text=uOrange!88!black] at ([yshift=.60cm]D.center) {\shortstack{KERNEL\\OPERATOR}};
  \node[eq] at ([yshift=-.04cm]D.center) {$\mathbf d\mapsto\mathbf z(\mathbf d)$};
  \node[sub] at ([yshift=-.66cm]D.center) {$(K_d+\lambda I)\Gamma=Y$};

  \node[stage,draw=uGold!88,fill=uGold!9,minimum width=3.30cm] (E) at (14.50,1.35) {};
  \node[head,text=uGold!72!black] at ([yshift=.60cm]E.center) {POSE + UQ};
  \node[eq] at ([yshift=.10cm]E.center) {$\mathbf z\mapsto(\mathbf r,\boldsymbol\kappa,R,\mathbf p_i)$};
  \node[sub] at ([yshift=-.50cm]E.center) {\shortstack{$SO(3)$ reconstruction\\+ uncertainty}};

  \draw[flow] (A.east)--(B.west);
  \draw[flow] (B.east)--(C.west);
  \draw[flow] (C.east)--(D.west);
  \draw[flow] (D.east)--(E.west);

  \draw[-Latex,uTeal!85!black,line width=.78pt,rounded corners=4pt]
    ([xshift=.72cm]E.south) -- ++(0,-.47) -| ([xshift=-.72cm]A.south);
  \node[font=\scriptsize\bfseries,text=black,fill=white,inner xsep=4pt,inner ysep=1pt]
    at (7.90,-.08) {kernel coverage $\rightarrow$ simulate $\rightarrow$ add data};

  \path[use as bounding box] (0.02,-.32) rectangle (16.18,2.55);
\end{tikzpicture}}
    \caption{Method overview. Mechanics-aware coordinates feed a learned nonstationary kernel metric; the kernel operator learns $\mathbf d\mapsto\mathbf z(\mathbf d)$, after which structured reconstruction recovers the TDCR pose and uncertainty. The same kernel geometry drives sequential acquisition.}
    \label{fig:method_overview}
\end{figure*}
Let $\mathbf d=(\boldsymbol\tau,L,\mathbf c,D,E,r_o)$ collect tendon loads, robot length, routing twist rates, routing radius, Young's modulus, and backbone radius. For each design, the mechanics solver returns a backbone $\mathbf r_{\mathbf d}(s)$ and material-frame orientation. We encode orientation through rotational strain $\boldsymbol\kappa_{\mathbf d}(s)$ and learn the parametric operator
\begin{equation}
\mathcal G:\mathbf d\longmapsto
\left(\frac{\mathbf r_{\mathbf d}(L\xi)}{L},
L\boldsymbol\kappa_{\mathbf d}(L\xi)\right),
\qquad \xi\in[0,1].
\label{eq:tdcr_operator}
\end{equation}
The input $\mathbf d$ is finite-dimensional, but the output of $\mathcal G$ is a pair of fields on $[0,1]$; throughout, \emph{operator learning} refers to learning this complete design-to-function map. We realize the map in two levels: (1) a spatial kernel converts each function-valued equilibrium to a finite coefficient vector $\mathbf z(\mathbf d)$, and (2) a design-space kernel learns the coefficient map $\mathbf d\mapsto\mathbf z(\mathbf d)$. Evaluating that second map at a new design therefore predicts an entire continuous robot configuration after decoding $\mathbf z$. The predicted strain is integrated on $SO(3)$ and combined with the known routing geometry. The same learned design kernel also supplies GP uncertainty and the geometry used for sequential simulator acquisition; see Figure~\ref{fig:method_overview}.

\subsection{Rotation-preserving output representation}
\label{sec:rotation_preserving_cosserat}
Let $\xi=s/L$ and $S=\{\xi_q\}_{q=1}^{n_s}\subset[0,1]$. We use the kernel-literature Mat\'ern $C^4$ kernel~\cite{Wendland2004,FasshauerMcCourt2015}
\begin{equation}
\phi(r)=(r^2+3r+3)e^{-r},\qquad
k_s(\xi,\xi')=\phi\!\left(\epsilon_s|\xi-\xi'|\right),
\label{eq:spatial_matern_kernel}
\end{equation}
with Gram matrix $(K_s)_{qj}=k_s(\xi_q,\xi_j)$. We choose $\epsilon_s$ from $\operatorname{cond}_1(K_s)= 10^{10}$ (making the kernel as flat as possible to increase approximation power)~\cite{FasshauerMcCourt2015} and reuse the same factorization for every equilibrium and both output fields.

We represent rotational strain rather than the entries of $R(s)$: the strain is an unconstrained Euclidean field, while its integration recovers a rotation without a posterior projection onto $SO(3)$. For equilibrium $\ell$, the two dimensionless output fields are
\begin{align}
\frac{\mathbf r_\ell(s)}{L_\ell}
&=\sum_{j=1}^{n_s}\mathbf a_{\ell,j}
 k_s\!\left(\frac{s}{L_\ell},\xi_j\right),
\label{eq:backbone_interpolant}\\
L_\ell\boldsymbol\kappa_\ell(s)
&=\sum_{j=1}^{n_s}\mathbf c_{\ell,j}
 k_s\!\left(\frac{s}{L_\ell},\xi_j\right).
\label{eq:kappa_interpolant}
\end{align}
Let $F_{\ell,r}$ and $F_{\ell,\kappa}$ contain the sampled dimensionless fields as rows. The coefficient matrices satisfy $K_sA_\ell=F_{\ell,r}$ and $K_sC_\ell=F_{\ell,\kappa}$, and one equilibrium is described by the coefficients
\begin{equation}
\mathbf z_\ell=
\begin{bmatrix}\operatorname{vec}(A_\ell)^T&\operatorname{vec}(C_\ell)^T\end{bmatrix}^T
\in\mathbb R^{6n_s}.
\label{eq:pose_coefficient_vector}
\end{equation}
Every training sample is thus one design paired with coefficients for two continuous fields; one spatial factorization produces the coefficient representation for the entire data set. Learning Eq.~\eqref{eq:tdcr_operator} therefore reduces to the finite-dimensional coefficient map $\mathbf d\mapsto\mathbf z(\mathbf d)$, while evaluation of the spatial kernel recovers the two fields at arbitrary arclength. We reconstruct orientation from the predicted strain through
\begin{equation}
R_\ell'(s)=R_\ell(s)[\boldsymbol\kappa_\ell(s)]_\times,
\qquad R_\ell(0)=I,
\label{eq:rotation_reconstruction_ivp}
\end{equation}
using a Lie-group integrator, which preserves $R_\ell(s)\in SO(3)$~\cite{IserlesMuntheKaasNorsettZanna2000}. Tendon $i$ then follows
\begin{equation}
\mathbf p_{i,\ell}(s)=\mathbf r_\ell(s)+R_\ell(s)\boldsymbol\rho_i(\mathbf d_\ell,s).
\label{eq:structured_tendon_reconstruction}
\end{equation}
The operator therefore learns one common pose, while tendon-specific geometry enters analytically through $\boldsymbol\rho_i$. The same spatial kernel also defines interpolation uncertainty. Its posterior mean is the interpolant above, and its covariance factor is
\begin{equation}
v_s(\xi,\xi')=k_s(\xi,\xi')-k_s(\xi,S)K_s^{-1}k_s(S,\xi'),
\label{eq:spatial_gp_covariance}
\end{equation}
the squared spatial power function in the noiseless case~\cite{Wendland2004,RasmussenWilliams2006}.

\subsection{Mechanics-aware input coordinates}
\label{sec:pose_design_coordinates}
The design kernel needs a notion of distance between robots. We define these distances through dimensionless combinations that enter bending, shear, and tendon routing. Let $I=\pi r_o^4/4$, $A=\pi r_o^2$, and $G=E/[2(1+\nu)]$. For tendon $i=1,\ldots,N_t$ and (equispaced) routing sample
$q=1,\ldots,n_\phi$, let $\eta_q\in\mathcal Q$ denote the normalized
arclength location and let $\alpha_i^0$ denote the base angle of tendon $i$. Define
\begin{equation}
b_i=\frac{\tau_iDL}{EI},\qquad
h_i=\operatorname{asinh}\!\left(\frac{\tau_i}{\sigma_hGA}\right),
\label{eq:dimensionless_load_coordinates}
\end{equation}
where $\sigma_h$ is a fixed shear scale. For normalized routing sites $\eta_q$ and tendon base angle $\alpha_i^0$, define
\begin{align}
\boldsymbol\psi_{i,q}
&=b_i\begin{bmatrix} \sin(\alpha_i^0+c_iL\eta_q) &
\cos(\alpha_i^0+c_iL\eta_q)
\end{bmatrix}^T,
\label{eq:loaded_routing_coordinates}\\
\mathbf m_q&=\sum_{i=1}^{N_t}\boldsymbol\psi_{i,q},\qquad
\mathbf g=\begin{bmatrix}D/L&r_o/L&D/r_o\end{bmatrix}^T.
\end{align}
Here $b_i$ measures the tendon loading relative to bending stiffness, $h_i$ separates the shear scale near zero, and $\boldsymbol\psi_{i,q}$ records how loading is distributed around the cross-section as the routing twists along the robot. The resultant $\mathbf m_q$ additionally keeps designs with similar net bending close even when individual tendons differ. We stack
\begin{equation}
\mathbf x(\mathbf d)=
\begin{bmatrix}
\mathbf b^T&\mathbf h^T&\operatorname{vec}(\boldsymbol\psi)^T&
\operatorname{vec}(\mathbf m)^T&\mathbf g^T
\end{bmatrix}^T,
\label{eq:pose_operator_coordinates}
\end{equation}
with dimension $d_x=2N_t+2N_tn_\phi+2n_\phi+3$ and five blocks $\mathcal B=\{b,h,\psi,m,g\}$. We keep these blocks separate because the nonstationary kernel below learns one local shape parameter for each mechanical contribution. Coordinates are standardized from the acquired designs. For the four-tendon robot, normalization moments are pooled over the cyclic $C_4$ orbit so quarter-turn rotations and the corresponding tendon relabelings receive identical normalization.

\subsection{Learned nonstationary GP operator}
\label{sec:kernel_operator_coefficients}
Let $\overline{\mathbf x}^{(a)}(\mathbf d)$ denote standardized block $a$. A shallow network supplies one positive local kernel shape per block:
\begin{equation}
\widetilde{\boldsymbol\epsilon}_\theta(\mathbf d)=
\epsilon_{\min}+\operatorname{softplus}\!\left(g_\theta(\overline{\mathbf x}(\mathbf d))\right).
\label{eq:shape_network}
\end{equation}
The network outputs only determine the local blockwise metric used when two designs are compared, allowing the relevant length scale to vary across design space without giving up the positive-definite GP construction. We impose $C_4$ invariance by geometric averaging,
\begin{equation}
\epsilon_a(\mathbf d)=\exp\!\left[
\frac1{|C_4|}\sum_{g\in C_4}\log\widetilde\epsilon_{\theta,a}(g\mathbf d)
\right].
\label{eq:c4_invariant_shapes}
\end{equation}
For two designs, set
\begin{align}
H_a(\mathbf d,\mathbf d')&=\tfrac12\left(\epsilon_a(\mathbf d)^{-2}+\epsilon_a(\mathbf d')^{-2}\right),\\
r_d(\mathbf d,\mathbf d')^2&=\sum_{a\in\mathcal B}
\frac{\|\overline{\mathbf x}^{(a)}(\mathbf d)-\overline{\mathbf x}^{(a)}(\mathbf d')\|_2^2}{H_a(\mathbf d,\mathbf d')},\\
q_d(\mathbf d,\mathbf d')&=\prod_{a\in\mathcal B}
\left[\epsilon_a(\mathbf d)\epsilon_a(\mathbf d')H_a(\mathbf d,\mathbf d')\right]^{-d_a/2}.
\end{align}
Here $\epsilon_a(\mathbf d)$ acts as a local inverse length scale for block $a$. The averaged metric $H_a$ compares two designs using both local scales while $q_d$ supplies the normalization required by the nonstationary covariance. The nonstationary construction~\cite{PaciorekSchervish2004}
\begin{equation}
k_d(\mathbf d,\mathbf d')=q_d(\mathbf d,\mathbf d')\phi\!\left(r_d(\mathbf d,\mathbf d')\right)
\label{eq:nonstationary_design_kernel}
\end{equation}
therefore remains positive definite as the learned metric varies. 
For acquired pairs $\{(\mathbf d_\ell,\mathbf z_\ell)\}_{\ell=1}^N$, we standardize the coefficients as $\mathbf y_\ell=(\mathbf z_\ell-\boldsymbol\mu_z)\oslash\boldsymbol\sigma_z$ and collect them in $Y$. With $(K_d)_{\ell m}=k_d(\mathbf d_\ell,\mathbf d_m)$, all output coefficients share one kernel-based operator learning system,
\begin{align}
(K_d&+\lambda I)\Gamma=Y,
\label{eq:operator_coefficient_solve}\\
\widehat{\mathbf y}(\mathbf d_*)&=k_d(\mathbf d_*,D)\Gamma,
\label{eq:operator_kernel_expansion}\\
\widehat{\mathbf z}(\mathbf d_*)&=
\boldsymbol\mu_z+\boldsymbol\sigma_z\odot\widehat{\mathbf y}(\mathbf d_*).
\end{align}
We enforce quarter-turn symmetry in the output through
\begin{equation}
\widehat{\mathbf z}_{C_4}(\mathbf d)=\frac1{|C_4|}
\sum_{g\in C_4}T_{g^{-1}}\widehat{\mathbf z}(g\mathbf d),
\label{eq:c4_averaged_prediction}
\end{equation}
where $T_g$ rotates backbone and strain coefficients. Interpreting $k_d$ as a GP covariance gives the common scalar covariance factor
\begin{equation}
v_d(\mathbf d,\mathbf d')=k_d(\mathbf d,\mathbf d')-
 k_d(\mathbf d,D)(K_d+\lambda I)^{-1}k_d(D,\mathbf d').
\label{eq:design_gp_covariance}
\end{equation}
Thus $v_d$ here measures uncertainty in the design-to-coefficient map, while $v_s$ in \eqref{eq:spatial_gp_covariance} measures interpolation uncertainty within each output function. These are distinct sources of uncertainty. 
Both are decoded through Eqs.~\eqref{eq:backbone_interpolant}--\eqref{eq:structured_tendon_reconstruction}; each sampled strain is integrated on $SO(3)$, so uncertainty propagation never produces an invalid orientation.

\subsection{Learning the design metric}
\label{sec:shape_network_training}
We learn $\theta$ in \eqref{eq:shape_network} through the kernel operator itself. At each update, acquired simulator equilibria are split into support and query sets $\mathcal I$ and $\mathcal J$. The support set defines a temporary kernel operator, which is evaluated on the query designs and decoded all the way back to physical tendon positions. Gradients therefore pass through the learned metric, kernel solve, spatial reconstruction, and $SO(3)$ decoder. If $P_j$ and $\widehat P_j(\theta)$ stack the true and predicted tendon positions for query robot $j$, we minimize
\begin{equation}
\mathcal L(\theta)=\frac1{|\mathcal J|}\sum_{j\in\mathcal J}
\frac{\|\widehat P_j(\theta)-P_j\|_F}{\|P_j\|_F}+\beta\mathcal R(\theta).
\label{eq:shape_training_objective}
\end{equation}
The regularizer prevents degenerate local shapes. An independent selection set chooses the checkpoint, after which the operator is refit on all acquired designs by solving \eqref{eq:operator_coefficient_solve}. This predictive objective selects the covariance through physical pose error.

\subsection{Sequential kernel-coverage acquisition}
\label{sec:active_kernel_sampling}
The learned kernel also defines coverage of the admissible design space. Let $\mathcal A$ be the acquired designs and $\mathcal C$ an input-only candidate pool. The corresponding design GP assigns a candidate the posterior variance
\begin{equation}
v_{\mathcal A}(\mathbf d)=k_d(\mathbf d,\mathbf d)-k_d(\mathbf d,\mathcal A)
(K_{\mathcal A}+\lambda I)^{-1}k_d(\mathcal A,\mathbf d).
\label{eq:active_gp_variance}
\end{equation}
Evaluating this quantity over a large candidate pool requires repeated solves with the acquired-set kernel. We therefore use a normalized kernel-coverage surrogate that depends only on pairwise kernel evaluations. For a candidate $\mathbf d$, define normalized similarity and coverage deficit
\begin{align}
\widetilde k_d(\mathbf d,\mathbf d')
&=\frac{k_d(\mathbf d,\mathbf d')}
{\sqrt{k_d(\mathbf d,\mathbf d)k_d(\mathbf d',\mathbf d')}},\\
\delta(\mathbf d;\mathcal A)
&=1-\max_{\mathbf a\in\mathcal A}\widetilde k_d(\mathbf d,\mathbf a).
\label{eq:kernel_coverage_deficit}
\end{align}
This is related to data-independent $P$-greedy selection~\cite{SantinHaasdonk2017}. The connection to GP variance is explicit in the interpolation limit. If $c=1-\delta$, then for $\lambda=0$,
\begin{equation}
\frac{v_{\mathcal A}(\mathbf d)}{k_d(\mathbf d,\mathbf d)}
\le 1-c^2=2\delta-\delta^2.
\label{eq:coverage_variance_bound}
\end{equation}
We greedily select the eligible candidate with largest $\delta$ and update coverage after each selection. Selection uses only candidate inputs and the current learned kernel; no candidate equilibrium is evaluated before selection. We simulate the selected designs, add their equilibria to $\mathcal A$, and refit the normalizer, metric, and operator before the next batch. Prescribed proportions of discrete actuation classes are maintained in every selected prefix.

\subsection{Scalable kernel operator learning}
\label{sec:askotch_scaling}
A direct solve of Eq.~\eqref{eq:operator_coefficient_solve} requires storing the $N\times N$ Gram matrix. We instead use ASkotch~\cite{askotch2026}, which performs block-coordinate updates with a low-rank Nystr\"om preconditioner while evaluating only required rows of the full kernel. With $\lambda_{\rm eff}=\max\{\lambda,\lambda_{\rm floor}\}$, the restricted residual for block $\mathcal I$ is
\begin{equation}
G_{\mathcal I}^{(t)}=(K_d)_{\mathcal I,:}\Gamma^{(t)}+
\lambda_{\rm eff}\Gamma_{\mathcal I,:}^{(t)}-Y_{\mathcal I,:}.
\label{eq:askotch_block_residual}
\end{equation}
All $d_z$ right-hand sides are solved jointly. We use block size $b=\max\{100,\lceil N/100\rceil\}$ and preconditioner rank $r=\min\{100,N\}$. The $r$ Nystr\"om landmarks are chosen independently of the sampled coordinate blocks, so the preconditioner rank is not tied to $b$. Crucially, this Nystr\"om approximation is only a preconditioner. Equation~\eqref{eq:askotch_block_residual} evaluates rows of the full learned kernel, so ASkotch approximates the solution of the original kernel system rather than replacing the operator by a rank-$r$ model. The fitted expansion still contains all $N$ acquired poses. To the best of our knowledge, this is the first application of ASkotch to kernel-based operator learning. In Section \ref{sec:results}, we compare ASkotch to using Cholesky factorizations for solving \eqref{eq:operator_coefficient_solve}.

\subsection{Sparse posterior uncertainty}
\label{sec:nystrom_uncertainty}
The mean and covariance use Nystr\"om approximations for different purposes. ASkotch uses one only to precondition the mean solve. For the posterior covariance, where the full Gram matrix would otherwise be required explicitly, we build a separate Fully Independent Training Conditional (FITC)-style Nystr\"om approximation~\cite{quinonero2005}. Its landmarks are sampled independently of the ASkotch preconditioner. Let $Z_u$ contain $r_u$ uncertainty landmarks and $K_d(Z_u,Z_u)+\epsilon_uI=L_uL_u^T$. Define whitened features and the FITC diagonal correction by
\begin{align}
\phi(d)&=L_u^{-1}K_d(Z_u,d),\\
\delta_i&=\max\{k_d(d_i,d_i)-\phi(d_i)^T\phi(d_i),0\}.
\label{eq:fitc_correction}
\end{align}
The low-rank term $\phi(d_i)^T\phi(d_i)$ captures variance represented by the landmark subspace; $\delta_i$ retains the remaining pointwise prior variance instead of discarding it as in a pure low-rank Nystr\"om covariance. With $\Phi_X$ containing $\phi(d_i)^T$ as rows,
\begin{equation}
W=\operatorname{diag}(\boldsymbol\delta+\lambda_u\mathbf1)^{-1},\qquad
A=I_{r_u}+\Phi_X^TW\Phi_X,
\label{eq:fitc_system}
\end{equation}
and the approximate design-space covariance is
\begin{equation}
\widehat v_d(d,d')=k_d(d,d')-\phi(d)^T\phi(d')+
\phi(d)^TA^{-1}\phi(d').
\label{eq:fitc_covariance}
\end{equation}
For our kernel, $k_d(d,d)=3$, so $\widehat v_d(d)=3-\|\phi(d)\|_2^2+\phi(d)^TA^{-1}\phi(d)$. The reduced system is accumulated in chunks, allowing all acquired designs to condition the approximate posterior without forming $K_d$.

To recover correlations among pose coefficients, let
\begin{align}
B_z=\frac1N\sum_{n=1}^N(\mathbf z_n-\overline{\mathbf z}&)
(\mathbf z_n-\overline{\mathbf z})^T,\\
\operatorname{Cov}[\mathbf z(\mathbf d),\mathbf z(\mathbf d')]
&\approx\widehat v_d(\mathbf d,\mathbf d')B_z.
\label{eq:multioutput_coefficient_covariance}
\end{align}
After applying the $C_4$ coefficient actions, denote the resulting query covariance by $\boldsymbol\Sigma_z(\mathbf d)$. We propagate position relative to the known backbone base using
\begin{equation}
\widetilde{\mathbf J}_{z,i}(\mathbf d,\xi)=
\mathbf J_{z,i}(\mathbf d,\xi)-\mathbf J_{z,\mathrm b}(\mathbf d,0).
\label{eq:anchored_coefficient_jacobian}
\end{equation}
Spatial interpolation uncertainty is combined with the empirical rotational-strain covariance into $\boldsymbol\Sigma_{\kappa,\mathrm{res}}$. We use a separable coefficient covariance $\widehat v_d B_z$ and first-order Jacobian propagation through the reconstruction. Assuming independence of the design-induced and spatial-residual terms,
\begin{equation}
\boldsymbol\Sigma_{p_i}(\mathbf d,\xi)=
\widetilde{\mathbf J}_{z,i}\boldsymbol\Sigma_z(\mathbf d)
\widetilde{\mathbf J}_{z,i}^T+
\mathbf J_{\kappa,i}\boldsymbol\Sigma_{\kappa,\mathrm{res}}
\mathbf J_{\kappa,i}^T\in\mathbb R^{3\times3}.
\label{eq:cartesian_position_covariance}
\end{equation}
Here $\mathbf J_{\kappa,i}$ propagates curvature residuals through the sequential $SO(3)$ integration and tendon-routing reconstruction. The first term propagates uncertainty in the design-to-coefficient operator; the second propagates the residual uncertainty of the spatial strain representation. Their sum defines the Cartesian ellipsoids and tubes. Base anchoring removes an artificial translational uncertainty at the known attachment point, so the design-induced position covariance vanishes there by construction.

\section{Results}
\label{sec:results}

\subsection{Preliminaries}
\label{sec:prelim}

\paragraph{Accuracy Metric}
We evaluate generalization using the relative $\ell_2$ error in the predicted
tendon trajectories. For test equilibrium $\ell$, let $P_\ell$ and
$\widehat P_\ell$ stack all ground-truth and predicted tendon positions,
respectively, using the same construction as in
Eq.~\eqref{eq:shape_training_objective}. The relative error is
\begin{equation}
    e_{\ell_2}^{(\ell)}
    =
    \frac{
        \left\|\widehat P_\ell-P_\ell\right\|_F
    }{
        \left\|P_\ell\right\|_F
    }.
    \label{eq:l2Error}
\end{equation}
Over a test set $\mathcal T$, we report the mean relative error
\begin{equation}
    \overline e_{\ell_2}
    =
    \frac{1}{|\mathcal T|}
    \sum_{\ell\in\mathcal T}e_{\ell_2}^{(\ell)}.
\end{equation}
For ease of interpretation, we also report the corresponding relative
percentage error, $(100 \times \overline e_{\ell_2})\%$.

\paragraph{Uncertainty}
Scalar uncertainty results report the pointwise latent posterior uncertainty,
using the FITC-style Nystr\"om approximation in Sec.~\ref{sec:nystrom_uncertainty} for ASkotch fits. Physical-space results report the combined design-space and spatial interpolation uncertainty from
Eq.~\eqref{eq:cartesian_position_covariance}. The resulting $3\times3$
Cartesian covariance matrices define the uncertainty ellipsoids and tubes,
which are anchored at the known backbone base.

\paragraph{Experimental setup}
We generate all equilibria with the Cosserat-rod simulator over the admissible
design space in Table~\ref{tab:design_space}. Training designs are acquired
either by IID sampling or by the sequential kernel-coverage procedure in
Sec.~\ref{sec:active_kernel_sampling}. We evaluate all models on a fixed,
independently sampled test set of 1000 equilibria. Active
acquisition begins from the same initial set as the IID baseline and selects
new designs from an input-only candidate pool in batches. The OOD study uses
separately generated query designs extending beyond the training distribution.
Timing experiments were performed on a machine with a RTX 4080 GPU with 16GB of RAM.
\begin{table}[t]
\centering
\caption{Admissible TDCR design and actuation space used to generate the
simulation data.}
\label{tab:design_space}
\footnotesize
\setlength{\tabcolsep}{5pt}
\begin{tabular}{lcc}
\hline
Parameter & Symbol & Range \\
\hline
Tendon load        & $\tau_i$ & [\,0,\,5\,]~N \\
Robot length       & $L$      & [\,0.10,\,0.35\,]~m \\
Routing twist rate & $c_i$    & [\,$-20$,\,20\,]~rad\,m$^{-1}$ \\
Routing radius     & $D$      & [\,0.005,\,0.020\,]~m \\
Young's modulus    & $E$      & [\,15.5,\,46.5\,]~GPa \\
Backbone radius    & $r_o$    & [\,0.0005,\,0.0015\,]~m \\
Number of tendons  & $N_t$    & 4 \\
Number of phase routing locations & $N_{\phi}$ & 4 \\
Number of spatial locations & $N_s$ & 42\\
\hline
\end{tabular}
\end{table}

\subsection{Generalization Errors}
\label{sec:accuracy}

\begin{figure}[htbp]
\centering
\includegraphics[width=0.49\textwidth]{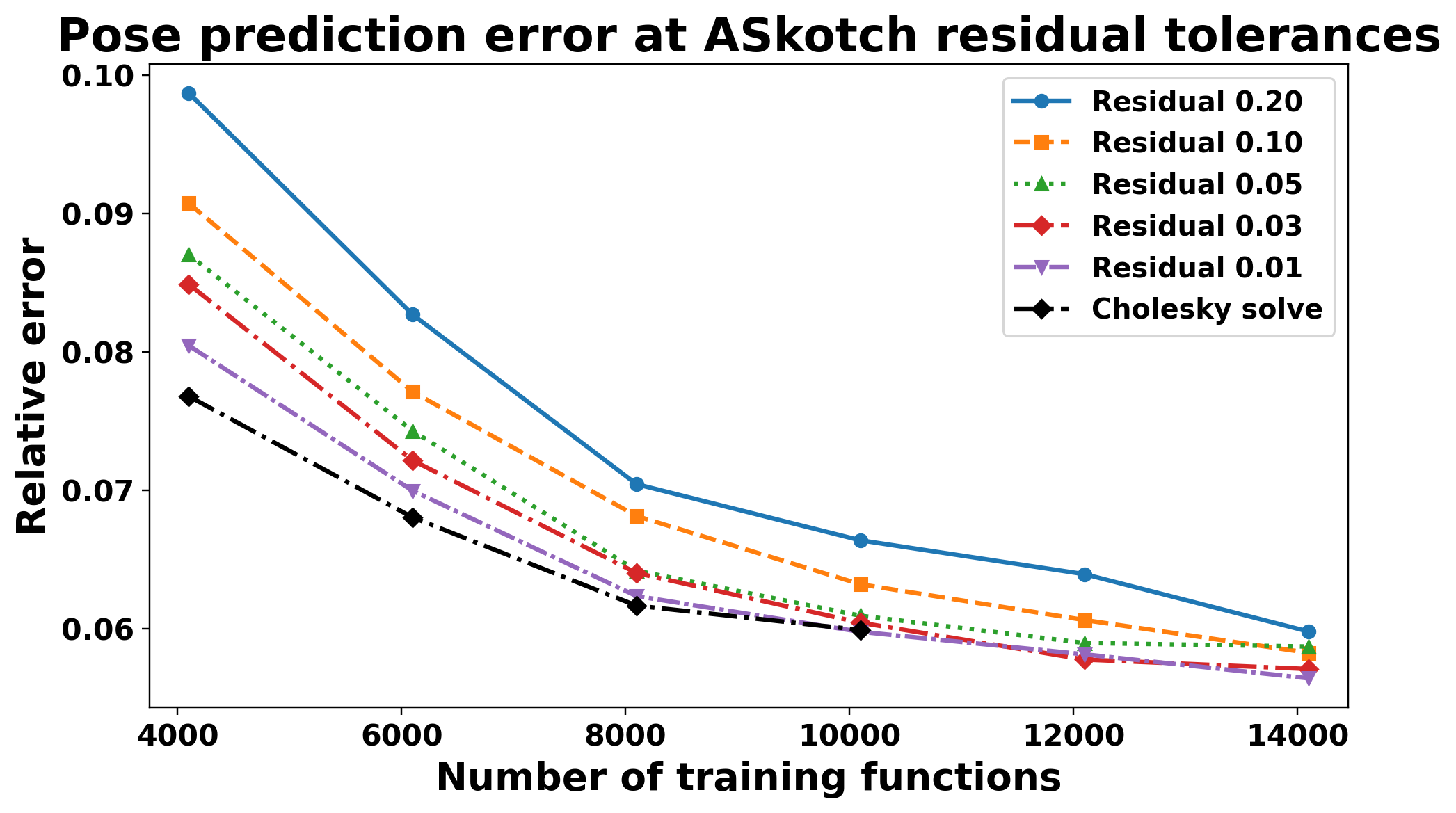}
\caption{Prediction accuracy for ASkotch under different target residuals, with the direct Cholesky solve as the baseline.}
\label{fig:residuals_target}
\end{figure}

\begin{figure}[htbp]
\centering
\includegraphics[width=0.49\textwidth]{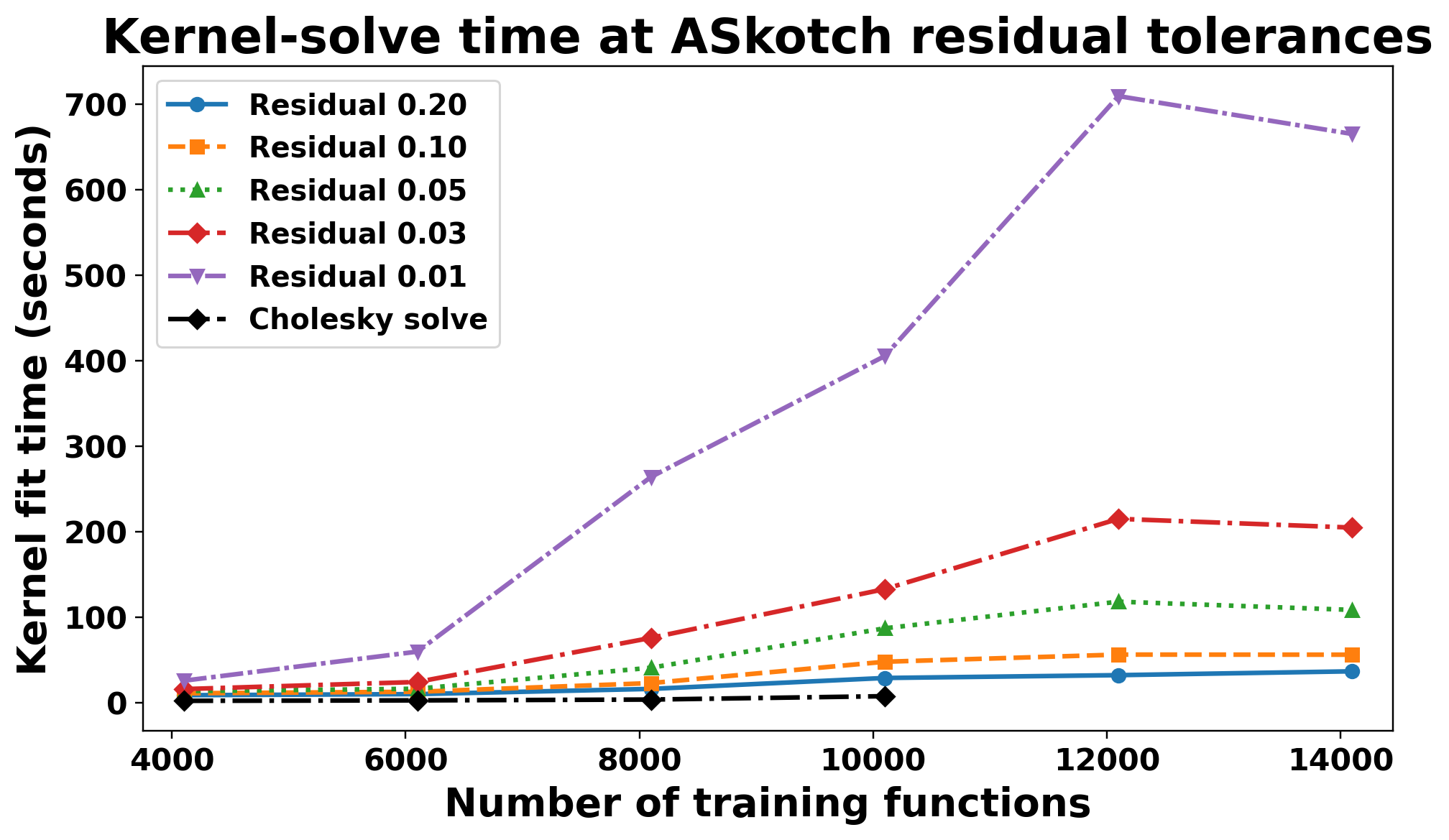}
\caption{Training time required to reach different ASkotch residual targets as the number of training functions increases. Note that the Cholesky solve was unable to be used past 10,000 training functions due to memory limitations.}
\label{fig:residuals_timing}
\end{figure}

First, we evaluate the ASkotch residual tolerance to identify a training regime that balances accuracy and computational cost. As shown in Fig.~\ref{fig:residuals_target}, decreasing the target residual improves prediction accuracy and approaches the direct solver baseline, with diminishing returns as the residual decreases. However, Fig.~\ref{fig:residuals_timing} shows a substantial increase in training time for tighter residual targets; clearly, ASkotch trades off memory with training times. At 12,000 training functions/designs, reaching a residual of $0.01$ requires more than twice the time required to reach a residual of $0.03$. For the remaining experiments, we therefore use a target residual of $0.03$, which provides a practical compromise between accuracy and training time.

\begin{figure}[htbp]
\centering
\includegraphics[width=0.49\textwidth]{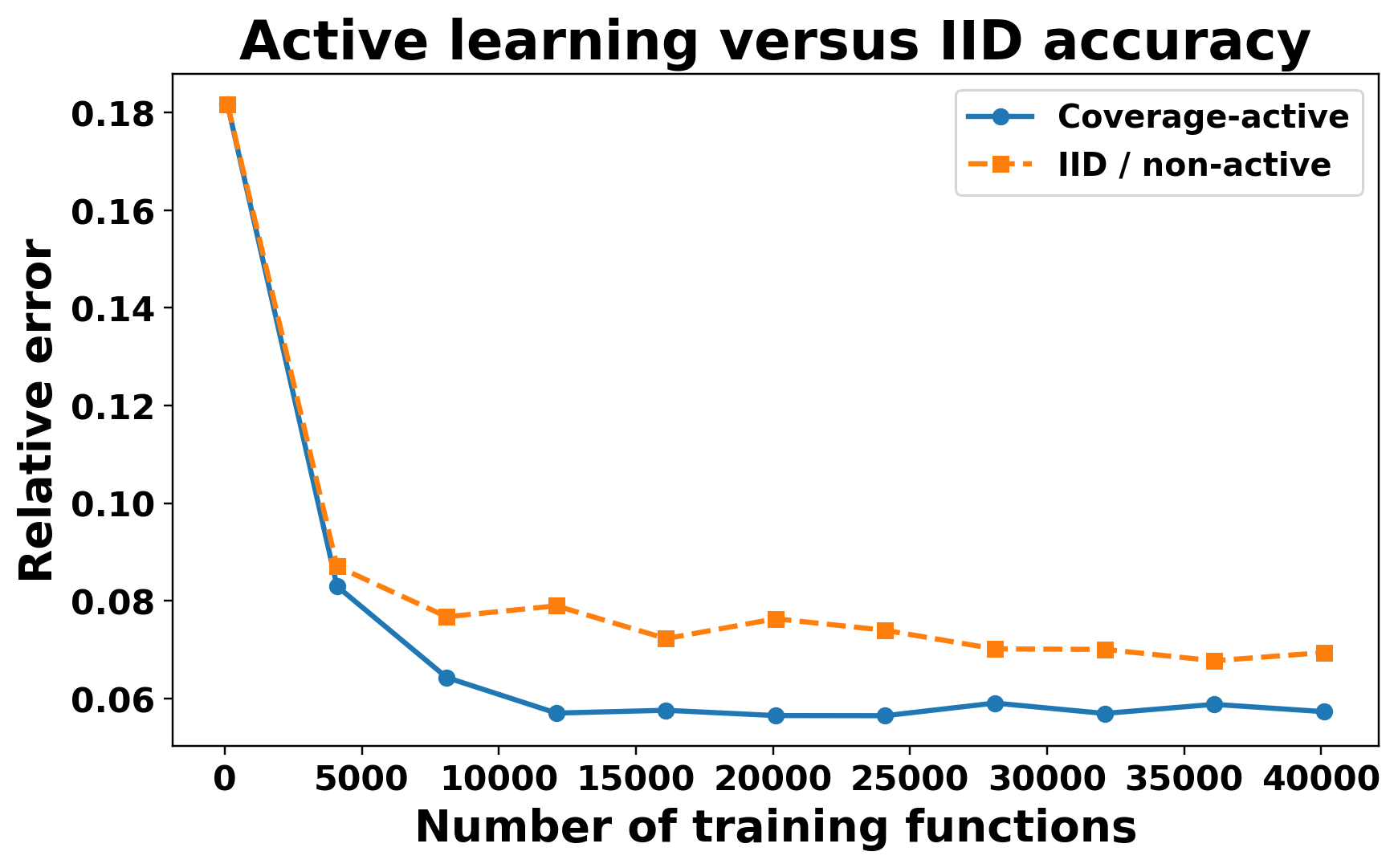}
\caption{Prediction accuracy for sequential kernel-coverage acquisition compared with IID sampling as the number of training functions increases.}
\label{fig:activeVpassive}
\end{figure}

Next, we evaluate the effect of sequential kernel-coverage acquisition on prediction accuracy. As shown in Fig.~\ref{fig:activeVpassive}, kernel-coverage acquisition reduces the mean relative error by approximately $15\%$ compared with independent and identically distributed (IID) sampling across the tested training-set sizes, consistently achieving lower errors throughout the evaluated range. Both acquisition strategies also achieve substantially lower errors than the DeepONet and FNO models reported in~\cite{FriedenFergusonKuntzShankar2026}, reducing the error by approximately $30\%$ with passive learning and $50\%$ with active learning compared to the pose DeepONet (the best model in that work). The two acquisition strategies use the same initial set of 100 training functions and therefore coincide at that data point. For the first 4000 functions, the difference between active and passive learning is relatively small. This is likely because the input space remains largely unexplored, such that IID sampling provides coverage comparable to that achieved by the active acquisition strategy. As the number of training functions increases, the active strategy increasingly targets underrepresented regions of the input space, leading to a larger performance advantage. At larger training-set sizes, however, this gap begins to narrow, potentially because the active set has already thoroughly explored the input space, while the IID samples progressively fill in the remaining gaps.

\subsection{Uncertainty}
\begin{figure}[htbp]
\centering
\includegraphics[width=0.49\textwidth]{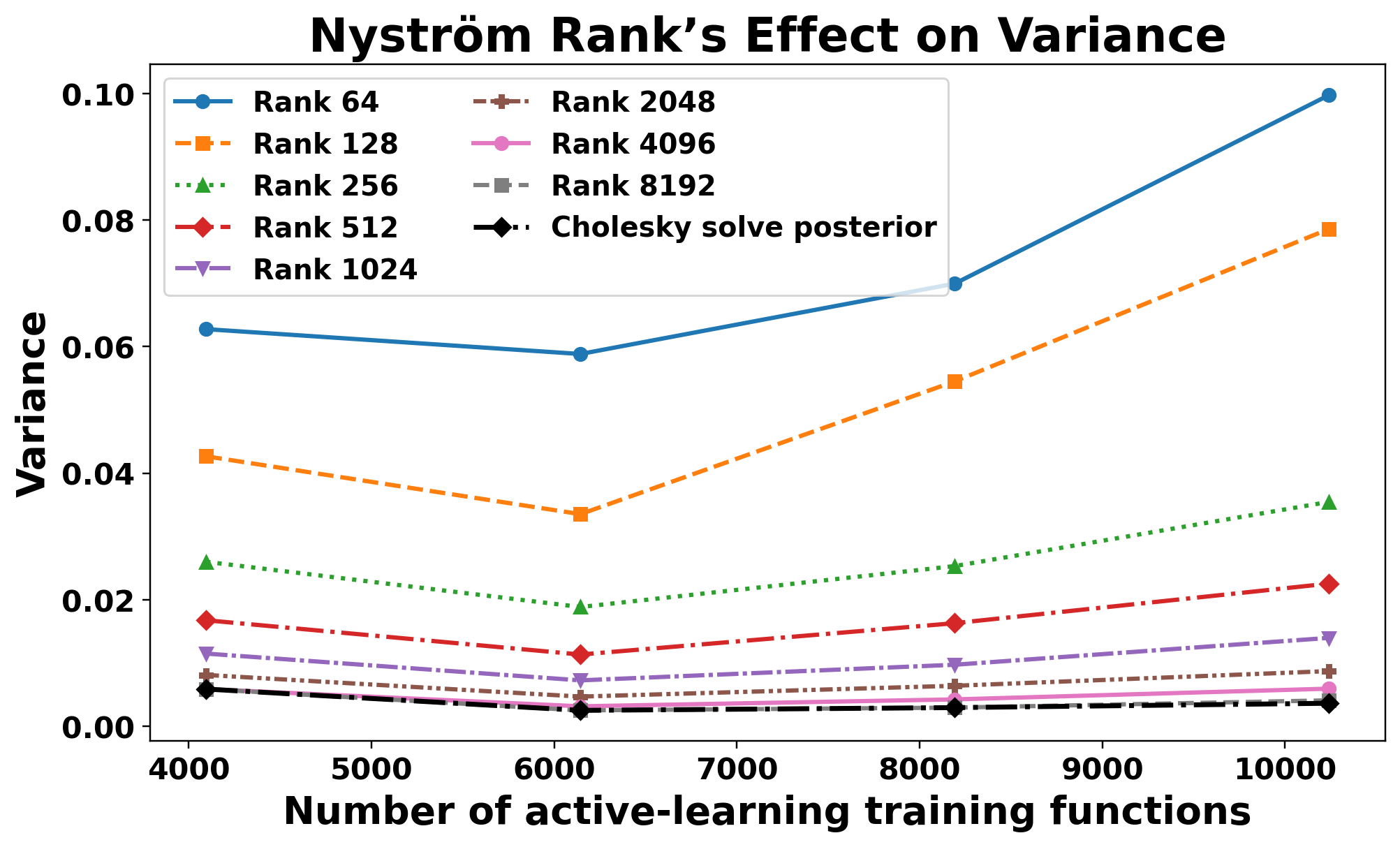}
\caption{Design-space posterior variance for different Nystr\"om ranks as the number of training functions increases, compared with the direct Cholesky solve baseline. These ranks were not used in training, only for estimation of the posterior variance}
\label{fig:Nyström_ranks}
\end{figure}

We next investigate how the Nystr\"om rank affects the approximate design-space posterior covariance relative to the direct Cholesky solve. As shown in Fig.~\ref{fig:Nyström_ranks}, increasing the Nystr\"om rank brings the approximate posterior variance progressively closer to the full-kernel result. When the requested rank reaches the number of training functions, the landmark set contains the complete training set and the Nystr\"om representation becomes full rank. In this limit, the FITC correction vanishes and the posterior recovers the direct Cholesky solved baseline, as observed for the 4096 and 8192 rank settings when their requested ranks are clipped to the corresponding training-set size.

For ranks below the training-set size, all acquired designs continue to condition the posterior, but the prior covariance is represented through a reduced Nystr\"om basis. Across all ranks and training-set sizes evaluated here, this approximation is conservative relative to the direct Cholesky solve posterior, yielding posterior variances greater than or equal to the full-rank result. For a fixed Nystr\"om rank, the discrepancy from the Cholesky  solved baseline generally increases as the number of training functions increases over the evaluated range. We therefore use a Nystr\"om rank of 8192 for the remaining experiments, which is the largest rank that can be used reliably within the available memory.

\begin{figure}[htbp]
\centering
\includegraphics[width=0.49\textwidth]{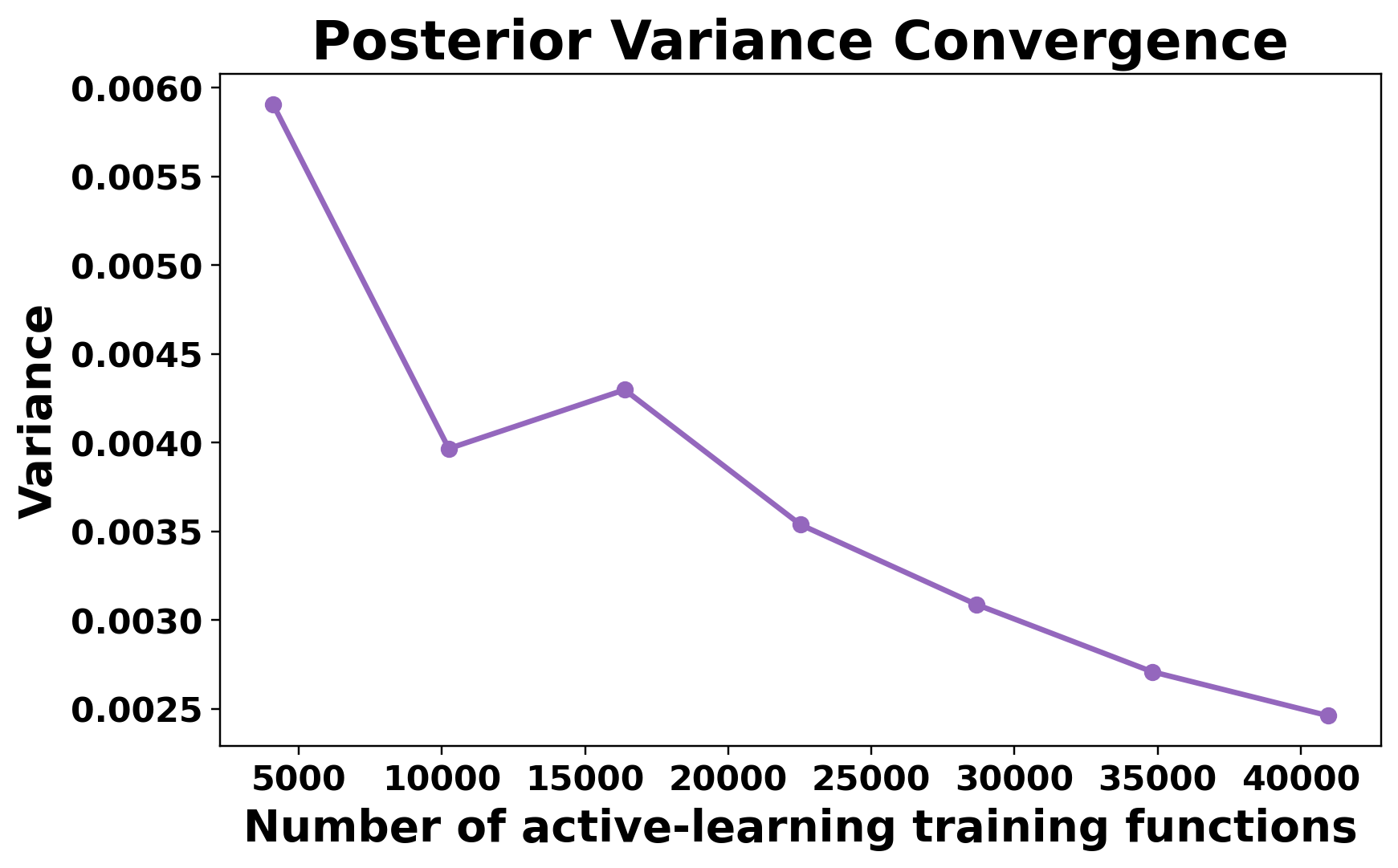}
\caption{Design-space posterior variance as the number of training functions increases.}
\label{fig:uncertainty_convergence}
\end{figure}

Next, we examine how the design-space posterior uncertainty changes over a larger range of training-set sizes. As shown in Fig.~\ref{fig:uncertainty_convergence}, increasing the number of training functions reduces the posterior variance, with the reduction gradually leveling off at larger training-set sizes. This behavior is consistent with progressively improved coverage of the design space as additional training functions are acquired.

The observed leveling may also be influenced by the fixed-rank Nystr\"om approximation. As the number of training functions increases while the Nystr\"om rank remains fixed, the low-rank representation must approximate an increasingly large kernel system. Consistent with the rank study above, this can make the approximate posterior increasingly conservative relative to the full-rank posterior over the evaluated range.

Next, we examine how the design-space posterior variance changes for increasingly out-of-distribution (OOD) designs. For a query design $\mathbf d$, we define its maximum similarity to the Nyström uncertainty landmarks $Z_u$ as
\begin{equation}
s_{\max}(\mathbf d)
=
\max_{\mathbf z\in Z_u}
k_d(\mathbf d,\mathbf z).
\end{equation}
Because the design-space kernel satisfies $k_d(\mathbf d,\mathbf d)=3$, we normalize the maximum landmark similarity to define the landmark-based OOD score
\begin{equation}
\operatorname{OOD}(\mathbf d)
=
1-\frac{s_{\max}(\mathbf d)}{3},
\end{equation}
where $\operatorname{OOD}(\mathbf d)\in[0,1]$. A score near zero indicates that the query design is highly similar to at least one landmark, whereas a score near one indicates low similarity to all landmarks.

\begin{figure}[htbp]
    \centering
    \includegraphics[width=0.49\textwidth]{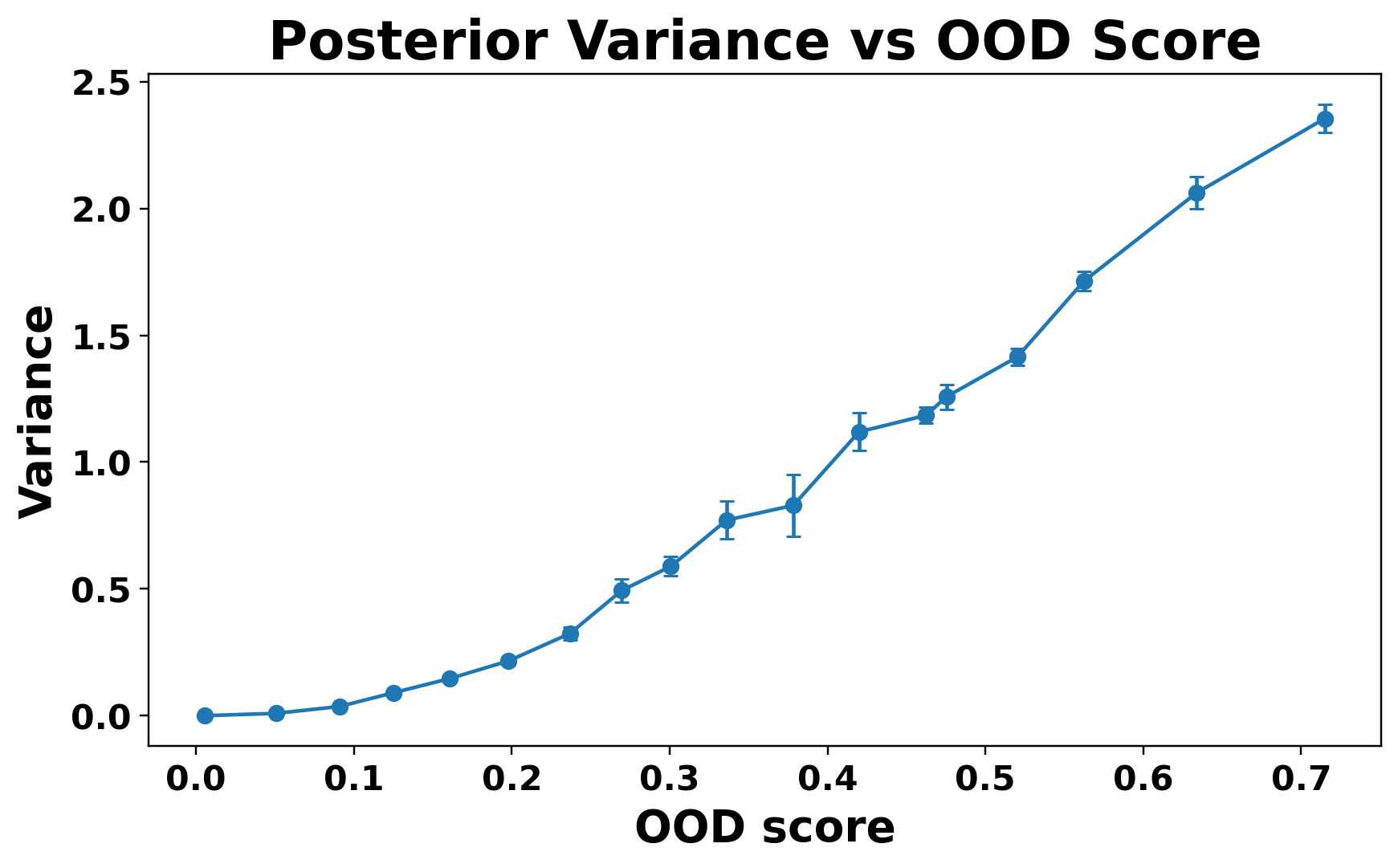}
    \caption{Design-space posterior variance as a function of the landmark-based
    OOD score.}
    \label{fig:uncertainty_OOD}
\end{figure}

We group the evaluated designs into bins according to their OOD scores and
report the mean posterior variance within each bin, with whiskers showing the
within-bin variation. As shown in Fig.~\ref{fig:uncertainty_OOD}, the posterior
variance initially increases gradually but rises more rapidly as the evaluated
designs move farther from the landmark-covered region of the design space. Clearly, our GP operator is able to provide useful information about its own uncertainty, an important feature for a practical surrogate model.

\subsection{Physical-Space Uncertainty}
\begin{figure}[htbp]
    \centering    \includegraphics[width=0.25\textwidth] 
    {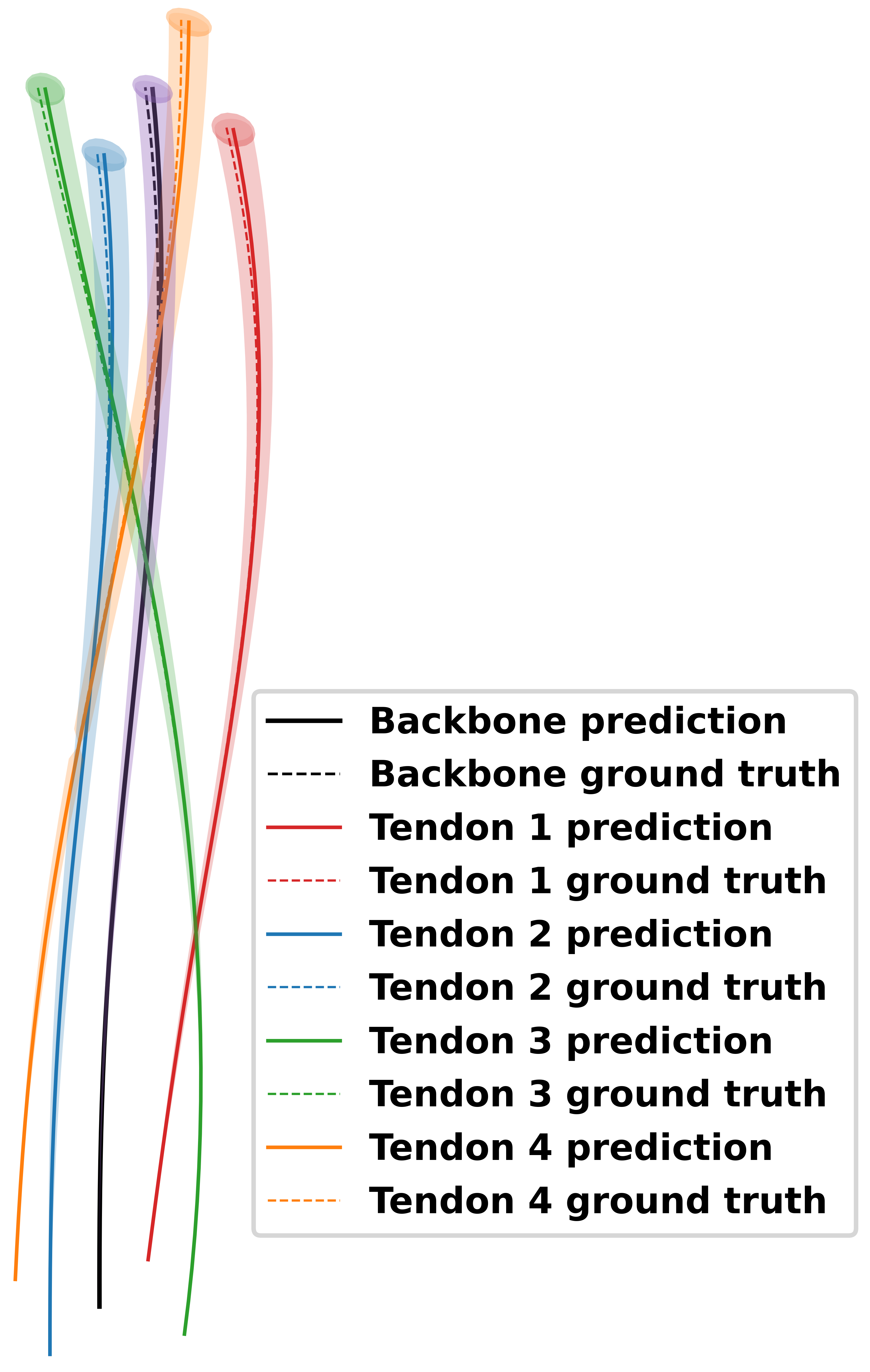}
    \caption{Base-anchored physical-space uncertainty for an example design,
    showing covariance tubes around the backbone and four tendons and the
    corresponding covariance ellipsoid at the robot tip.}
    \label{fig:tubes}
\end{figure}

We next examine how the design-space and spatial interpolation uncertainties
are propagated into Cartesian space. Figure~\ref{fig:tubes} shows the resulting
uncertainty tubes for the backbone and four tendons of an example design. The
uncertainty is smallest at the known backbone base. In particular, the
design-induced positional covariance vanishes there by construction because
of the base-anchored Jacobian in
Eq.~\eqref{eq:anchored_coefficient_jacobian}.

For this example, the uncertainty tubes generally widen toward the robot tip.
This behavior reflects the increasing sensitivity of distal positions to
perturbations in the predicted pose coefficients, together with the cumulative
effect of spatial curvature residuals propagated through the sequential
$SO(3)$ reconstruction and tendon-routing map. The shape and orientation of
each tube cross-section are determined by the corresponding full Cartesian
covariance matrix, including correlations between coordinate directions. 

\section{Conclusion}
\label{sec:conclusion}
We developed a probabilistic kernel/GP operator surrogate for equilibrium prediction across a broad four-tendon TDCR design space. The representation learns backbone position and rotational strain, reconstructs orientation directly on $SO(3)$, and introduces tendon geometry analytically after reconstruction. Mechanics-aware dimensionless coordinates and a nonstationary $C_4$-symmetric Mat\'ern kernel define the design-space model. The same kernel supports sequential simulator acquisition, while ASkotch and a separate Nystr\"om-FITC approximation make the mean and posterior uncertainty practical for larger training sets. The resulting uncertainty can be propagated through the structured reconstruction to Cartesian covariance tubes.

Our experiments show a consistent accuracy gain from kernel-coverage acquisition over IID sampling and convergence of the sparse posterior toward the direct GP posterior as Nystr\"om rank increases. Posterior variance also rises as designs move away from the landmark-covered region, providing a useful indicator of extrapolation. These results make the surrogate suitable not only for fast prediction but also for deciding where additional mechanics solves are most valuable. This work thus takes significant steps toward a learned model that is accurate for the spectrum of designs rather than specific to one, enabling faster optimization, control, and planning for these robots.  Future work will test calibration against physical TDCR data and extend the construction beyond the four-tendon symmetry used here.

\bibliographystyle{IEEEtran}
\bibliography{refs}

@article{Webster2010ConstantCurvature,
  author  = {Robert J. Webster III and Bryan A. Jones},
  title   = {Design and Kinematic Modeling of Constant Curvature Continuum Robots: A Review},
  journal = {The International Journal of Robotics Research},
  year    = {2010},
  volume  = {29},
  number  = {13},
  pages   = {1661--1683},
  doi     = {10.1177/0278364910368147}
}

@article{Burgner2015_TRO,
  author  = {Jessica Burgner-Kahrs and D. Caleb Rucker and Howie Choset},
  title   = {Continuum Robots for Medical Applications: A Survey},
  journal = {IEEE Transactions on Robotics},
  year    = {2015},
  volume  = {31},
  number  = {6},
  pages   = {1261--1280},
  doi     = {10.1109/TRO.2015.2489500}
}

@article{Dupont2022_ProcIEEE,
  author  = {Pierre E. Dupont and Nabil Simaan and Howie Choset and Caleb Rucker},
  title   = {Continuum Robots for Medical Interventions},
  journal = {Proceedings of the IEEE},
  year    = {2022},
  volume  = {110},
  number  = {7},
  pages   = {847--870},
  doi     = {10.1109/JPROC.2022.3141338}
}

@article{Rucker2011StaticsTendons,
  author  = {D. Caleb Rucker and Robert J. Webster III},
  title   = {Statics and Dynamics of Continuum Robots With General Tendon Routing and External Loading},
  journal = {IEEE Transactions on Robotics},
  year    = {2011},
  volume  = {27},
  number  = {6},
  pages   = {1033--1044},
  doi     = {10.1109/TRO.2011.2160461}
}

@article{Rao2021HowToModelTDCR,
  author  = {Priyanka Rao and Quentin Peyron and Sven Lilge and Jessica Burgner-Kahrs},
  title   = {How to Model Tendon-Driven Continuum Robots and Benchmark Modelling Performance},
  journal = {Frontiers in Robotics and AI},
  year    = {2021},
  volume  = {7},
  pages   = {630245},
  doi     = {10.3389/frobt.2020.630245}
}

@article{Till2019RealtimeCosserat,
  author  = {Jonathan Till and Carlo Aloi and D. Caleb Rucker},
  title   = {Real-Time Dynamics of Soft and Continuum Robots Based on Cosserat Rod Models},
  journal = {The International Journal of Robotics Research},
  year    = {2019},
  volume  = {38},
  number  = {12--13},
  pages   = {1575--1592},
  doi     = {10.1177/0278364919842269}
}

@misc{FriedenFergusonKuntzShankar2026,
  author        = {Branden Frieden and James M. Ferguson and Alan Kuntz and Varun Shankar},
  title         = {Neural Operators for Design-Space Surrogate Modeling of Tendon-Actuated Continuum Robots},
  year          = {2026},
  eprint        = {2605.19104},
  archivePrefix = {arXiv},
  primaryClass  = {cs.RO},
  doi           = {10.48550/arXiv.2605.19104}
}

@article{BatlleDarcyHosseiniOwhadi2024,
  author  = {Pau Batlle and Matthieu Darcy and Bamdad Hosseini and Houman Owhadi},
  title   = {Kernel Methods are Competitive for Operator Learning},
  journal = {Journal of Computational Physics},
  volume  = {496},
  pages   = {112549},
  year    = {2024},
  doi     = {10.1016/j.jcp.2023.112549}
}

@book{RasmussenWilliams2006,
  author    = {Carl Edward Rasmussen and Christopher K. I. Williams},
  title     = {Gaussian Processes for Machine Learning},
  publisher = {MIT Press},
  address   = {Cambridge, MA},
  year      = {2006}
}

@book{Wendland2004,
  author    = {Holger Wendland},
  title     = {Scattered Data Approximation},
  series    = {Cambridge Monographs on Applied and Computational Mathematics},
  volume    = {17},
  publisher = {Cambridge University Press},
  year      = {2004},
  doi       = {10.1017/CBO9780511617539}
}

@book{FasshauerMcCourt2015,
  author    = {Gregory E. Fasshauer and Michael J. McCourt},
  title     = {Kernel-Based Approximation Methods Using {MATLAB}},
  series    = {Interdisciplinary Mathematical Sciences},
  volume    = {19},
  publisher = {World Scientific},
  year      = {2015}
}

@article{IserlesMuntheKaasNorsettZanna2000,
  author  = {Arieh Iserles and Hans Z. Munthe-Kaas and Syvert P. N{\o}rsett and Antonella Zanna},
  title   = {Lie-Group Methods},
  journal = {Acta Numerica},
  volume  = {9},
  pages   = {215--365},
  year    = {2000},
  doi     = {10.1017/S0962492900002154}
}

@inproceedings{PaciorekSchervish2004,
  author    = {Christopher J. Paciorek and Mark J. Schervish},
  title     = {Nonstationary Covariance Functions for Gaussian Process Regression},
  booktitle = {Advances in Neural Information Processing Systems 16},
  editor    = {Sebastian Thrun and Lawrence K. Saul and Bernhard Sch{\"o}lkopf},
  publisher = {MIT Press},
  year      = {2004}
}

@article{SantinHaasdonk2017,
  author  = {Gabriele Santin and Bernard Haasdonk},
  title   = {Convergence Rate of the Data-Independent {$P$}-Greedy Algorithm in Kernel-Based Approximation},
  journal = {Dolomites Research Notes on Approximation},
  volume  = {10},
  pages   = {68--78},
  year    = {2017},
  doi     = {10.14658/pupj-drna-2017-Special_Issue-9}
}

@article{askotch2026,
  author  = {Pratik Rathore and Zachary Frangella and Jiaming Yang and Micha{\l} Derezi{\'n}ski and Madeleine Udell},
  title   = {Have {ASkotch}: A Neat Solution for Large-Scale Kernel Ridge Regression},
  journal = {Journal of Machine Learning Research},
  volume  = {27},
  number  = {162},
  pages   = {1--48},
  year    = {2026}
}

@article{quinonero2005,
  author  = {Joaqu{\'i}n Qui{\~n}onero-Candela and Carl Edward Rasmussen},
  title   = {A Unifying View of Sparse Approximate Gaussian Process Regression},
  journal = {Journal of Machine Learning Research},
  year    = {2005},
  volume  = {6},
  pages   = {1939--1959}
}

@book{Antman2005Problems,
  author    = {Stuart S. Antman},
  title     = {Nonlinear Problems of Elasticity},
  edition   = {2},
  series    = {Applied Mathematical Sciences},
  volume    = {107},
  publisher = {Springer},
  address   = {New York},
  year      = {2005},
  doi       = {10.1007/0-387-27649-1}
}

@article{DeisenrothFoxRasmussen2015,
  author  = {Marc Peter Deisenroth and Dieter Fox and Carl Edward Rasmussen},
  title   = {Gaussian Processes for Data-Efficient Learning in Robotics and Control},
  journal = {IEEE Transactions on Pattern Analysis and Machine Intelligence},
  year    = {2015},
  volume  = {37},
  number  = {2},
  pages   = {408--423},
  doi     = {10.1109/TPAMI.2013.218}
}

@article{LilgeBarfootBurgnerKahrs2022,
  author  = {Sven Lilge and Timothy D. Barfoot and Jessica Burgner-Kahrs},
  title   = {Continuum Robot State Estimation Using Gaussian Process Regression on {$SE(3)$}},
  journal = {The International Journal of Robotics Research},
  year    = {2022},
  volume  = {41},
  number  = {13--14},
  pages   = {1099--1120},
  doi     = {10.1177/02783649221128843}
}

@article{RelanoMunozMonje2023,
  author  = {Carlos Rela{\~n}o and Javier Mu{\~n}oz and Concepci{\'o}n A. Monje},
  title   = {Gaussian Process Regression for Forward and Inverse Kinematics of a Soft Robotic Arm},
  journal = {Engineering Applications of Artificial Intelligence},
  year    = {2023},
  volume  = {126},
  pages   = {107174},
  doi     = {10.1016/j.engappai.2023.107174}
}

@article{MoWeiKongZhaiLiuSun2025,
  author  = {Hangjie Mo and Ruofeng Wei and Xiaowen Kong and Yujia Zhai and Yunhui Liu and Dong Sun},
  title   = {Data-Efficient Learning Control of Continuum Robots in Constrained Environments},
  journal = {IEEE Transactions on Automation Science and Engineering},
  year    = {2025},
  volume  = {22},
  pages   = {984--995},
  doi     = {10.1109/TASE.2024.3357816}
}

@article{Tang2025HapticNO,
  author  = {Xiongfeng Tang and Tao Yang and Jian Wang and Guowei Han and Qiran Sun and Hanmin Peng and Zhijun Sun},
  title   = {Neural Operator-Based Haptic Sensing Along Slender Rods},
  journal = {International Journal of Mechanical Sciences},
  year    = {2025},
  volume  = {304},
  pages   = {110699},
  doi     = {10.1016/j.ijmecsci.2025.110699}
}

@misc{VeilFlaschelKuhlDellaSantina2026,
  author        = {Carina Veil and Moritz Flaschel and Ellen Kuhl and Cosimo Della Santina},
  title         = {Infinite-Dimensional Closed-Loop Inverse Kinematics for Soft Robots via Neural Operators},
  year          = {2026},
  eprint        = {2602.18655},
  archivePrefix = {arXiv},
  primaryClass  = {cs.RO},
  doi           = {10.48550/arXiv.2602.18655}
}

@article{MoraYousefpourHosseinmardiOwhadiBostanabad2025,
  author  = {Carlos Mora and Amin Yousefpour and Shirin Hosseinmardi and Houman Owhadi and Ramin Bostanabad},
  title   = {Operator Learning with Gaussian Processes},
  journal = {Computer Methods in Applied Mechanics and Engineering},
  year    = {2025},
  volume  = {434},
  pages   = {117581},
  doi     = {10.1016/j.cma.2024.117581}
}

@article{KumarNayekChakraborty2025,
  author  = {Sawan Kumar and Rajdip Nayek and Souvik Chakraborty},
  title   = {Towards Gaussian Process for Operator Learning: An Uncertainty Aware Resolution Independent Operator Learning Algorithm for Computational Mechanics},
  journal = {Computer Methods in Applied Mechanics and Engineering},
  year    = {2025},
  volume  = {435},
  pages   = {117664},
  doi     = {10.1016/j.cma.2024.117664}
}

@article{YangKissasPerdikaris2022,
  author  = {Yang, Yibo and Kissas, Georgios and Perdikaris, Paris},
  title   = {Scalable uncertainty quantification for deep operator networks using randomized priors},
  journal = {Computer Methods in Applied Mechanics and Engineering},
  volume  = {399},
  pages   = {115399},
  year    = {2022},
  doi     = {10.1016/j.cma.2022.115399}
}

@article{LinMoyaZhang2023,
  author  = {Lin, Guang and Moya, Christian and Zhang, Zecheng},
  title   = {{B-DeepONet}: An enhanced Bayesian {DeepONet} for solving noisy parametric {PDEs} using accelerated replica exchange {SGLD}},
  journal = {Journal of Computational Physics},
  volume  = {473},
  pages   = {111713},
  year    = {2023},
  doi     = {10.1016/j.jcp.2022.111713}
}

@article{GargChakraborty2023,
  author  = {Garg, Shailesh and Chakraborty, Souvik},
  title   = {{VB-DeepONet}: A Bayesian operator learning framework for uncertainty quantification},
  journal = {Engineering Applications of Artificial Intelligence},
  volume  = {118},
  pages   = {105685},
  year    = {2023},
  doi     = {10.1016/j.engappai.2022.105685}
}

@article{MaPittAzizzadenesheliAnandkumar2024,
  author  = {Ma, Ziqi and Pitt, David and Azizzadenesheli, Kamyar and Anandkumar, Anima},
  title   = {Calibrated Uncertainty Quantification for Operator Learning via Conformal Prediction},
  journal = {Transactions on Machine Learning Research},
  year    = {2024}
}

@article{MoyaMollaaliZhangLuLin2025,
  author  = {Moya, Christian and Mollaali, Amirhossein and Zhang, Zecheng and Lu, Lu and Lin, Guang},
  title   = {Conformalized-{DeepONet}: A distribution-free framework for uncertainty quantification in deep operator networks},
  journal = {Physica D: Nonlinear Phenomena},
  volume  = {471},
  pages   = {134418},
  year    = {2025},
  doi     = {10.1016/j.physd.2024.134418}
}

@inproceedings{MagnaniPfortnerWeberHennig2025,
  author    = {Magnani, Emilia and Pf{\"o}rtner, Marvin and Weber, Tobias and Hennig, Philipp},
  title     = {Linearization Turns Neural Operators into Function-Valued {Gaussian} Processes},
  booktitle = {Proceedings of the 42nd International Conference on Machine Learning},
  series    = {Proceedings of Machine Learning Research},
  volume    = {267},
  pages     = {42421--42452},
  year      = {2025},
  publisher = {PMLR}
}

@article{KumarNayekChakraborty2024,
  author  = {Kumar, Sawan and Nayek, Rajdip and Chakraborty, Souvik},
  title   = {Neural Operator induced Gaussian Process framework for probabilistic solution of parametric partial differential equations},
  journal = {Computer Methods in Applied Mechanics and Engineering},
  volume  = {431},
  pages   = {117265},
  year    = {2024},
  doi     = {10.1016/j.cma.2024.117265}
}
\end{document}